\documentclass[11pt]{article}
\usepackage{acl2015}
\usepackage{times}
\usepackage{url}
\usepackage{latexsym}
\usepackage{multirow}
\usepackage{rotating}
\usepackage{amsmath}
\usepackage{amsthm}
\usepackage{amssymb}
\usepackage{colortbl}
\usepackage{booktabs}
\usepackage{tabularx}
\usepackage{paralist}
\usepackage{graphicx}
\usepackage{lpic}
\usepackage[utf8]{inputenc}
\usepackage[small]{caption}
\usepackage{titlesec}

\title{AutoExtend: Extending Word Embeddings to Embeddings for Synsets and Lexemes}

\author{Sascha Rothe and Hinrich Sch\"{u}tze\\
Center for Information \& Language Processing\\
University of Munich\\
  {\tt sascha@cis.lmu.de} 
}

\date{}

\def\figref#1{Figure~\ref{fig:#1}}
\def\figlabel#1{\label{fig:#1}\label{p:#1}}

\def\tabref#1{Table~\ref{tab:#1}}
\def\tablabel#1{\label{tab:#1}\label{p:#1}}

\def\secref#1{Section~\ref{sec:#1}}
\def\seclabel#1{\label{sec:#1}\label{p:#1}}
\def\eqref#1{Eq.~\ref{eqn:#1}}
\def\eqrefn#1{\ref{eqn:#1}}
\def\eqsref#1#2{Eqs.~\ref{eqn:#1}-\ref{eqn:#2}}
\def\eqlabel#1{\label{eqn:#1}}

\long\def\symbolfootnote[#1]#2{\begingroup%
\def\thefootnote{\fnsymbol{footnote}}\footnote[#1]{#2}\endgroup}

\def\dnrm#1{\mbox{$_{\hbox{\scriptsize #1}}$}}

\begin{document}

\maketitle

\begin{abstract}
We present \textit{AutoExtend}, a system to learn embeddings
for synsets and lexemes. It is flexible in that it can take
any word embeddings as input and does not need an additional
training corpus. The synset/lexeme embeddings obtained live
in the same vector space as the word embeddings. A sparse
tensor formalization guarantees efficiency and
parallelizability. We use WordNet as a lexical resource, but
AutoExtend can be easily applied to other resources like
Freebase. AutoExtend achieves
state-of-the-art performance on word similarity and
word sense disambiguation
tasks.\end{abstract}

\section{Introduction}\seclabel{intro}
Unsupervised methods for word embeddings (also called
“distributed word representations”) have become popular in
natural language processing (NLP). These methods only need
very large corpora as input to create sparse
representations (e.g., based on local collocations) and project them into
a lower dimensional dense vector space. Examples for word
embeddings are SENNA  \cite{collobert2008unified}, the
hierarchical log-bilinear model 
\cite{mnih2009scalable}, word2vec 
\cite{mikolov2013distributed} and GloVe 
\cite{pennington2014glove}.  However, there are many
other resources that are undoubtedly useful in NLP,
including lexical resources like WordNet and Wiktionary and
knowledge bases like Wikipedia and Freebase. We will simply
call these \emph{resources} in the rest of the paper.  Our
goal is to enrich these valuable resources with embeddings
for those data types that are not words; e.g., we want to
enrich WordNet with embeddings for synsets and lexemes. A
\textit{synset} is a set of synonyms that are
interchangeable in some context. A \textit{lexeme} pairs a
particular spelling or pronunciation with a particular
meaning, i.e., a lexeme is a conjunction of a word and a
synset.  Our premise is that many NLP applications will
benefit if the non-word data types of resources -- e.g.,
synsets in WordNet -- are also available as embeddings. For
example, in machine translation, enriching and improving
translation dictionaries
(cf.\ \newcite{mikolov2013exploiting}) would benefit from
these embeddings because they would enable us to create an
enriched dictionary for word senses. Generally, our premise
is that the arguments for the utility of embeddings for word
forms should carry over to the utility of embeddings for
other data types like synsets in WordNet.

The insight underlying the method we propose is that
\emph{the constraints of a resource can be formalized as
 constraints on embeddings and then allow us to extend word
 embeddings to embeddings of other data types like
 synsets.} For example, the hyponymy relation in WordNet
can be formalized as such a constraint. 

The advantage of our approach is that it decouples
embedding learning from the extension of embeddings to
non-word data types in a resource. If somebody comes up with
a better way of learning embeddings, these embeddings become
immediately usable for resources. And we do not rely on any
specific properties of embeddings that make them
usable in some resources, but not in others.

An alternative to our approach is to train embeddings on
annotated text, e.g., to train synset embeddings on corpora
annotated with synsets. However, successful embedding learning generally requires very large
corpora and sense labeling is too expensive to produce corpora of such a
size.

Another alternative to our approach is to add up all word
embedding vectors related to a particular node in a
resource; e.g., to create the synset vector of \textit{lawsuit}
in WordNet, we can add the word vectors of the three words that
are part of the synset (\textit{lawsuit}, \textit{suit}, \textit{case}). We
will call this approach \emph{naive} and use it as a
baseline (S$\dnrm{naive}$ in \tabref{ims}).

We will focus on WordNet \cite{fellbaum1998} in this paper, but 
our method -- based on a formalization that exploits the constraints of a resource for
extending embeddings from words to other data types -- is
broadly applicable to other resources including Wikipedia
and Freebase.

A word in WordNet can be viewed as a composition of several
lexemes. Lexemes from different words together can form a
synset. When a synset is given, it can be decomposed into
its lexemes. And these lexemes then join to form words.
These observations are the basis for the formalization of
the constraints encoded in WordNet that will be presented in
the next section: \emph{we view words as the sum of their lexemes
and, analogously, synsets as the sum of their lexemes.}

Another motivation for our formalization stems from the
analogy calculus developed by
\newcite{mikolov2013efficient}, which can be viewed as a
group theory formalization of word relations:
we have a set of elements (our vectors) and an operation
(addition) satisfying the properties of a mathematical group, in particular,
associativity and invertibility. For example, you can take the vector of
\textit{king}, subtract the vector of \textit{man} and add
the vector of \textit{woman} to get a vector near
\textit{queen}. In other words, you remove the properties of
\textit{man} and add the properties of \textit{woman}. We
can also see the vector of \textit{king} as the sum of the
vector of \textit{man} and the vector of a gender-neutral
ruler. The next thing to notice is that this does not only
work for words that combine several \emph{properties}, but also
for words that combine several \emph{senses}. The vector of
\textit{suit} can be seen as the sum of a vector representing
\textit{lawsuit} and a vector representing
\textit{business suit}. AutoExtend is designed to take word
vectors as input and unravel the word vectors to the vectors of
their lexemes. The lexeme vectors will then give us the synset
vectors.

The main contributions of this paper are: (i) We present
AutoExtend, a flexible method that extends word embeddings
to embeddings of synsets and lexemes. AutoExtend is
completely general in that it can be used for any set of embeddings
and for any resource that imposes constraints of a certain
type on the relationship between words and other data
types. (ii) We show that AutoExtend achieves 
state-of-the-art word similarity
and word sense disambiguation (WSD)
performance. (iii) We publish the AutoExtend code for
extending word embeddings to other data types, the lexeme
and synset embeddings and the software to replicate our WSD
evaluation.

This paper is structured as follows. \secref{model} introduces the model, first as a general tensor formulation then as a matrix formulation making additional assumptions. In \secref{data}, we describe data, experiments and evaluation. We analyze AutoExtend in \secref{analysis} and give a short summary on how to extend our method to other resources in \secref{extending}. \secref{related} discusses related work.

\section{Model}\seclabel{model}
We are looking for a model that extends standard embeddings
for words to embeddings for the other two data types in
WordNet: synsets and lexemes. We want all three data types
-- words, lexemes, synsets -- to live in the same embedding
space.

The basic premise of our model is: (i) words are sums of their
lexemes and (ii) synsets are sums of their lexemes. 
We refer to these two premises as \emph{synset constraints}.
For example, the embedding of the word \textit{bloom} is a sum
of the embeddings of its two lexemes \textit{bloom(organ)}
and \textit{bloom(period)}; and the embedding of the synset
\textit{flower-bloom-blossom(organ)} is a sum of the
embeddings of its three lexemes \textit{flower(organ)},
\textit{bloom(organ)} and \textit{blossom(organ)}.

The synset constraints can be argued to be the simplest possible
relationship between the three WordNet data types. They can
also be motivated by the way many embeddings are learned from
corpora -- for example, the counts in vector space models
are additive, supporting the view of words as the sum of
their senses. The same assumption is frequently made; for
example, it underlies the group theory formalization of analogy 
discussed in \secref{intro}.

We denote word vectors as $w^{(i)} \in \mathbb{R}^n$, synset vectors as
$s^{(j)} \in \mathbb{R}^n$, and lexeme vectors as $l^{(i,j)} \in \mathbb{R}^n$. $l^{(i,j)}$ is
that lexeme of word $w^{(i)}$ that is a member of synset $s^{(j)}$.
We set lexeme vectors $l^{(i,j)}$ that do not exist
to zero. For example, the non-existing lexeme \textit{flower(truck)}
is set to zero. We can then formalize our premise that the two
constraints (i) and (ii) hold as follows:
\begin{align}
w^{(i)} = \sum_{j} l^{(i,j)} \eqlabel{word_lexeme}\\
s^{(j)} = \sum_{i} l^{(i,j)} \eqlabel{lexeme_synset}
\end{align}
These two equations are underspecified. We therefore 
introduce the matrix $E^{(i,j)} \in \mathbb{R}^{n \times n}$:
\begin{equation} \eqlabel{weights_encoding}
l^{(i,j)} = E^{(i,j)} w^{(i)}
\end{equation}
We make the assumption that the dimensions
in \eqref{weights_encoding} are independent of each other, i.e., 
$E^{(i,j)}$ is a diagonal matrix. Our motivation for this
assumption is:
(i) This makes the computation technically feasible by
significantly reducing the number of parameters and by
supporting parallelism. (ii) Treating word
embeddings on a per-dimension basis is a frequent
design choice (e.g., \newcite{KalchbrennerACL2014}).
Note that we allow $E^{(i,j)}<0$ and in
general the distribution weights
for each dimension (diagonal entries of $E^{(i,j)}$) will be different. 
Our assumption can be interpreted as word $w^{(i)}$
distributing its embedding activations to its lexemes on each
dimension separately. Therefore, \eqsref{word_lexeme}{lexeme_synset} can be written as follows:
\begin{align}
w^{(i)} = \sum_{j} E^{(i,j)} w^{(i)} \eqlabel{word_lexeme_weight}\\
s^{(j)} = \sum_{i} E^{(i,j)} w^{(i)} \eqlabel{lexeme_synset_weight}
\end{align}
Note that from \eqref{word_lexeme_weight} it directly follows that:
\begin{equation} \eqlabel{weights_encoding_normalized}
\sum_{j} E^{(i,j)} = I_n \quad \forall i
\end{equation}
with $I_n$ being the identity matrix.

Let $W$ be a $|W| \times n$ matrix where $n$ is the dimensionality of
the embedding space, $|W|$ is the number of words and each row $w^{(i)}$ is a word embedding;
and let $S$ be a $|S| \times n$ matrix where $|S|$ is
the number of synsets and each row $s^{(j)}$ is a synset
embedding. $W$ and $S$ can be interpreted as linear maps
and a mapping between them is given by the rank 4
tensor $\mathbf{E} \in \mathbb{R}^{|S| \times n \times |W| \times
 n}$. We can then write
\eqref{lexeme_synset_weight} as a tensor product:
\begin{align}
S = \mathbf{E} \otimes W
\end{align}
while \eqref{weights_encoding_normalized} states, that
\begin{equation}\eqlabel{e_normalized}
\sum_{j} \mathbf{E}_{j,d_2}^{i,d_1} = 1 \quad \forall i, d_1, d_2
\end{equation}
Additionally, there is no interaction between
different dimensions, so $\mathbf{E}_{j,d_2}^{i,d_1} = 0$ if $d_1 \neq
d_2$. In other words, we are creating the tensor by stacking the diagonal matrices $E^{(i,j)}$ over $i$
and $j$. Another sparsity arises from the fact that many
lexemes do not exist: $\mathbf{E}_{j,d_2}^{i,d_1} = 0$ if $l^{(i,j)} = 0$;
i.e., $l^{(i,j)}\neq 0$ only if word $i$ has a lexeme that is a member of
synset $j$. To summarize the sparsity:
\begin{equation}\eqlabel{e_sparse}
\mathbf{E}_{j,d_2}^{i,d_1} = 0 \Leftarrow d_1 \neq d_2 \vee l^{(i,j)} = 0
\end{equation}

\subsection{Learning}
We adopt an autoencoding framework to learn embeddings for
lexemes and synsets. To this end, we view the tensor
equation $S = \mathbf{E} \otimes W$ as the encoding part of the
autoencoder: the synsets are the encoding of the words.
We define a corresponding decoding part 
that decodes the synsets into words as follows:
\begin{equation}
s^{(j)} = \sum_{i} \overline{l}^{(i,j)}
\eqlabel{synset_lexeme}, \ \ \ \
\overline{w}^{(i)} = \sum_{j} \overline{l}^{(i,j)} 
\end{equation}
In analogy to $E^{(i,j)}$, we introduce the diagonal matrix $D^{(j,i)}$:
\begin{equation}
\overline{l}^{(i,j)} = D^{(j,i)} s^{(j)} \eqlabel{weights_decoding}
\end{equation}
In this case, it is the synset that distributes itself to its lexemes. We can then rewrite \eqref{synset_lexeme} to:
\begin{equation}
s^{(j)} = \sum_{i} D^{(j,i)} s^{(j)},\ \overline{w}^{(i)} = \sum_{j} D^{(j,i)} s^{(j)}
\end{equation}
and we also get the equivalent of \eqref{weights_encoding_normalized} for $D^{(j,i)}$:
\begin{equation}\eqlabel{weights_decoding_normalized}
\sum_{i} D^{(j,i)} = I_n \quad \forall j
\end{equation}
and in tensor notation:
\begin{align}
\overline{W} = \mathbf{D} \otimes S
\end{align}
Normalization and sparseness properties for the decoding part
are analogous to the encoding part:
\begin{equation}\eqlabel{d_normalized}
\sum_{i} \mathbf{D}_{i,d_1}^{j,d_2} = 1 \quad \forall j, d_1, d_2
\end{equation}
\begin{equation}\eqlabel{d_sparse}
\mathbf{D}_{i,d_1}^{j,d_2} = 0 \Leftarrow d_1 \neq d_2 \vee l^{(i,j)} = 0
\end{equation}
We can state the learning objective of the autoencoder as follows:
\begin{equation}\eqlabel{objective_tensor}
\underset{\mathbf{E},\mathbf{D}}{\operatorname{argmin}} \|\mathbf{D} \otimes \mathbf{E} \otimes W - W\|
\end{equation}
under the conditions \eqref{e_normalized}, \eqrefn{e_sparse}, \eqrefn{d_normalized} and \eqrefn{d_sparse}.

Now we have an autoencoder where input and output layers are
the word embeddings and the hidden layer represents the
synset vectors. A simplified version is shown in
\figref{autoencoder}. The tensors $\mathbf{E}$ and $\mathbf{D}$ have to be
learned. They are rank 4 tensors of size ${\approx}10^{15}$.
However, we already discussed that they are very
sparse, for two reasons: (i) We make the assumption that there is no interaction between
dimensions. (ii) There are only few interactions between
words and synsets (only when a lexeme exists). In practice, there
are only ${\approx}10^7$ elements to learn, which is
technically feasible.

\subsection{Matrix formalization}
Based on the assumption that
each dimension is fully independent from other
dimensions, a separate autoencoder for each
dimension can be created and trained in parallel. Let $W \in
\mathbb{R}^{|W| \times n}$ be a matrix where each row is a
word embedding and $w^{(d)} = W_{\cdot,d} $ the $d$-th column of $W$, i.e.,
a vector that holds the $d$-th dimension of each word
vector. In the same way, $s^{(d)} = S_{\cdot,d}$ holds the $d$-th
dimension of each synset vector and $E^{(d)} = \mathbf{E}^{\cdot,d}_{\cdot,d} \in \mathbb{R}^{|S| \times |W|}$. We can write $S =
\mathbf{E} \otimes W$ as:
\begin{equation}\eqlabel{synsets_dimensionwise}
s^{(d)} = E^{(d)} w^{(d)} \quad \forall d
\end{equation}
with $E^{(d)}_{i,j} = 0$ if $l^{(i,j)} = 0$. The decoding equation
$\overline{W} = \mathbf{D} \otimes S$ takes this form:
\begin{equation}
\overline{w}^{(d)} = D^{(d)} s^{(d)} \quad \forall d
\end{equation}
where $D^{(d)} = \mathbf{D}^{\cdot,d}_{\cdot,d} \in \mathbb{R}^{|W| \times |S|}$ and $D^{(d)}_{j,i} = 0$ if $l^{(i,j)} = 0$. So $E$ and $D$ are symmetric in terms of non-zero elements. The learning objective becomes:
\begin{equation}\eqlabel{objective_matrix}
\underset{E^{(d)},D^{(d)}}{\operatorname{argmin}} \|D^{(d)} E^{(d)} w^{(d)} - w^{(d)}\| \quad \forall d
\end{equation}

\begin{figure*}
\centering
\includegraphics[width=16cm]{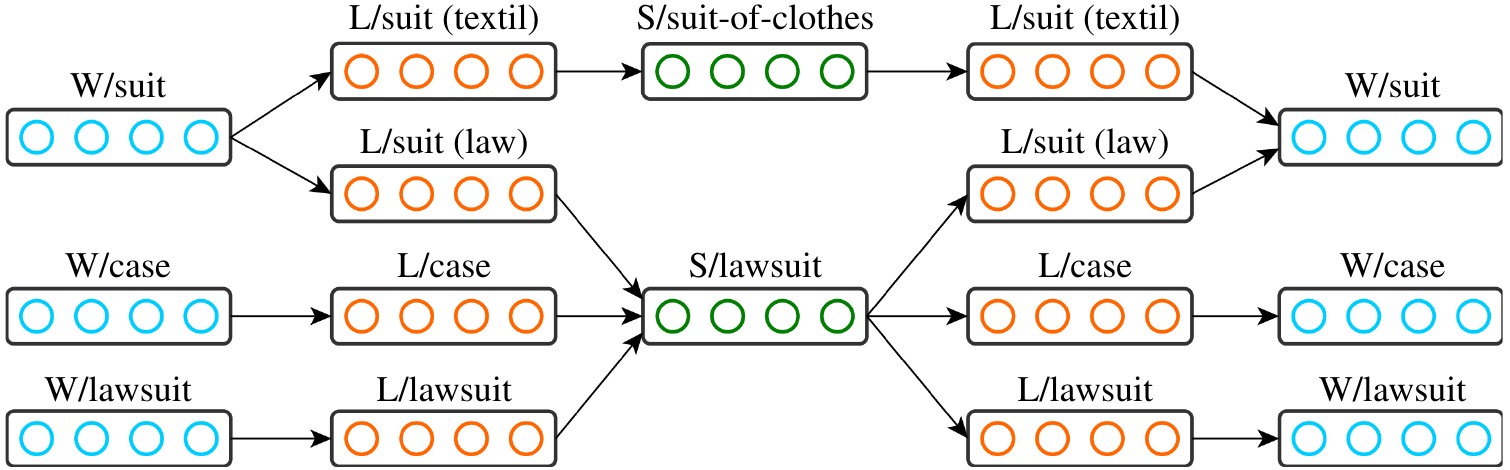}
\caption{A small subgraph of WordNet. The circles are
  intended to show four different embedding
  dimensions. These dimensions are treated as
  independent. The synset constraints align the input and
  the output layer. The lexeme constraints align the second and fourth layers.}
\figlabel{autoencoder}
\end{figure*}

\subsection{Lexeme embeddings}
The hidden layer $S$ of the autoencoder 
gives us synset embeddings. The lexeme embeddings are
defined when transitioning from $W$ to $S$, or more explicitly
by:
\begin{equation}
l^{(i,j)} = E^{(i,j)} w^{(i)}
\end{equation}
However, there is also a second lexeme embedding in AutoExtend when transitioning form $S$ to $\overline{W}$:
\begin{equation}
\overline{l}^{(i,j)} = D^{(j,i)} s^{(j)}
\end{equation}
Aligning these two representations seems natural, so we impose
the following \emph{lexeme constraints}:
\begin{equation}
\underset{E^{(i,j)},D^{(j,i)}}{\operatorname{argmin}} \left\|
E^{(i,j)} w^{(i)} - D^{(j,i)} s^{(j)}
\right\| \quad \forall i,j
\end{equation}
This can also be expressed dimension-wise. The matrix formulation is given by:
\begin{equation}\eqlabel{objective_lexeme}
\resizebox{\hsize}{!}{$\underset{E^{(d)},D^{(d)}}{\operatorname{argmin}} \left\| E^{(d)} \operatorname{diag}(w^{(d)}) - \left(D^{(d)} \operatorname{diag}(s^{(d)})\right)^T \right\| \forall d$}
\end{equation}
with $\operatorname{diag}(x)$ being a square matrix having
$x$ on the main diagonal and vector $s^{(d)}$ defined by
\eqref{synsets_dimensionwise}. While we try to align the 
embeddings, there are still two different
lexeme embeddings. In all experiments reported in \secref{analysis}
we will use the average of both embeddings and in \secref{analysis} we will
analyze the weighting in more detail.

\subsection{WN relations}\seclabel{synset_relations}
\begin{table}
\centering
\begin{tabular}{l|rrrr}
& noun & verb & adj & adv \\\hline
hypernymy & 84,505 & 13,256 & 0 & 0 \\\hline
antonymy & 2,154 & 1,093 & 4,024 & 712 \\\hline
similarity & 0 & 0 & 21,434 & 0 \\\hline
verb group & 0 & 1,744 & 0 & 0 
\end{tabular}
\caption{\# of WN relations by part-of-speech}
\tablabel{relations}
\end{table}
Some WordNet synsets contain only a single word (lexeme). The autoencoder learns based on the synset constraints, i.e., lexemes being shared by different synsets (and also words); thus, it is difficult to learn good embeddings for single-lexeme synsets. To remedy this problem, we impose the constraint that \emph{synsets related by WordNet (WN) relations should have similar embeddings}. \tabref{relations} shows relations
we used. WN relations are entered in a new matrix $R \in \mathbb{R}^{r \times |S|}$, where $r$ is the number of WN relation tuples. For each relation tuple, i.e., row in $R$, we set the columns corresponding to the first and second synset to $1$ and $-1$, respectively. The values of $R$ are not updated during training. We use a squared error function and
$0$ as target value. This forces the system to find similar
values for related synsets. Formally, the \emph{WN relation constraints} are:
\begin{equation}\eqlabel{objective_relations}
\underset{E^{(d)}}{\operatorname{argmin}} \|R E^{(d)} w^{(d)}\| \quad \forall d
\end{equation}

\subsection{Implementation}\seclabel{implementation} 
Our training objective is minimization of the sum of synset
constraints (\eqref{objective_matrix}), weighted by
$\alpha$, the lexeme constraints (\eqref{objective_lexeme}),
weighted by $\beta$, and the WN relation constraints
(\eqref{objective_relations}), weighted by $1-\alpha-\beta$.

The training objective cannot be solved analytically because it is subject to constraints \eqref{e_normalized}, \eqref{e_sparse}, \eqref{d_normalized} and \eqref{d_sparse}. We therefore use backpropagation.
We do not use regularization since we found that all learned weights are in $[-2, 2]$.

AutoExtend is implemented in MATLAB. We run $1000$
iterations of gradient descent. On an Intel Xeon CPU E7-8857
v2 3.00GHz, one iteration on one dimension takes less than a
minute because the gradient computation ignores zero entries
in the matrix.

\subsection{Column normalization}\seclabel{column_normalized}
Our model is based on the premise that a word is the sum of
its lexemes (\eqref{word_lexeme}). From the definition of
$E^{(i,j)}$, we derived that $\mathbf{E} \in \mathbb{R}^{|S|
  \times n \times |W| \times n}$ is normalized over the
first dimension (\eqref{e_normalized}). So $E^{(d)} \in
\mathbb{R}^{|S| \times |W|}$ is also normalized over the
first dimension. In other words, $E^{(d)}$ is a column
normalized matrix.  Another premise of the model is that a
synset is the sum of its lexemes. Therefore, $D^{(d)}$ is
also column normalized. A simple way to implement this is
to start
the computation with column normalized matrices and
normalize them again after each iteration as long as the
error function still decreases. When the error function
starts increasing, we stop normalizing the matrices and
continue with a normal gradient descent. This respects that
while $E^{(d)}$ and $D^{(d)}$ should be column normalized in
theory, there are a lot of practical issues that prevent
this, e.g., OOV words.

\begin{table}[b]
\centering
\begin{tabular}{l|rr}
 & WordNet & $\cap$ word2vec\\\hline
words & 147,478 & 54,570 \\\hline
synsets & 117,791 & 73,844\\\hline
lexemes & 207,272 & 106,167
\end{tabular}
\caption{\# of items in WordNet and after intersection with word2vec vectors}
\tablabel{corpus_size}
\end{table}

\section{Data, experiments and evaluation}\seclabel{data}\seclabel{evaluation}
We downloaded 300-dimensional embeddings for 3,000,000 words and
phrases trained on Google News, a corpus of ${\approx}10^{11}$ tokens, using word2vec CBOW
\cite{mikolov2013distributed}. Many 
words in the word2vec vocabulary
are not in WordNet, e.g.,
inflected forms (\textit{cars}) and proper nouns
(\textit{Tony Blair}). Conversely, many
WordNet lemmas are not in the word2vec vocabulary, e.g., \textit{42} 
(digits were converted to 0). This
results in a number of empty synsets (see
\tabref{corpus_size}). Note however that
AutoExtend can produce embeddings for empty synsets
because we use WN relation constraints in addition to synset and lexeme constraints.

We run AutoExtend on the word2vec vectors. As we do not know anything about a suitable weighting for the three different constraints, we set $\alpha = \beta = 0.33$. Our main goal is to produce compatible embeddings for lexemes and synsets. Thus, we can compute nearest neighbors across all three data types as shown in \figref{nearest_neighbors}.
\begin{figure}[tb]
\small
\begin{tabular}{p{0.46\textwidth}}
\multicolumn{1}{c}{\normalsize nearest neighbors of W/suit}\\
S/suit (businessman), L/suit (businessman),
L/accomodate, S/suit (be acceptable), L/suit (be
acceptable), L/lawsuit, W/lawsuit, S/suit (playing card),
L/suit (playing card), S/suit (petition), S/lawsuit,
W/countersuit, W/complaint, W/counterclaim\\\hline
\multicolumn{1}{c}{\rule{0pt}{3ex} \normalsize nearest neighbors of W/lawsuit}\\
L/lawsuit, S/lawsuit, S/countersuit,
L/countersuit, W/countersuit, W/suit, W/counterclaim,
S/counterclaim (n), L/counterclaim (n), S/counterclaim (v),
L/counterclaim (v), W/sue, S/sue (n), L/sue (n)\\\hline
\multicolumn{1}{c}{\rule{0pt}{3ex} \normalsize nearest neighbors of S/suit-of-clothes}\\
L/suit-of-clothes, S/zoot-suit,
L/zoot-suit, W/zoot-suit, S/garment, L/garment, S/dress,
S/trousers, L/pinstripe, L/shirt, W/tuxedo, W/gabardine,
W/tux, W/pinstripe
\end{tabular}
\caption{Five nearest word (W/), lexeme (L/) and synset (S/) neighbors for three items, ordered by cosine}
\figlabel{nearest_neighbors}
\end{figure}

We evaluate the embeddings on WSD and on similarity
performance. Our results depend directly on the quality of
the underlying word embeddings, in our case word2vec embeddings. We would expect even better evaluation
results as word representation learning methods
improve. Using a new and improved set of underlying
embeddings is simple: it is a simple switch of the input
file that contains the word embeddings.

\subsection{Word Sense Disambiguation}\seclabel{wsd}
For WSD we use the shared tasks of Senseval-2 \cite{kilgarriff2001english} and Senseval-3 \cite{mihalcea2004itri} and a system named IMS \cite{zhong2010makes}. Senseval-2 contains 139, Senseval-3 57 different words. They provide 8,611, respectively 8,022 training instances and 4,328, respectively 3,944 test instances. For the system, we use the same setting as in the original paper. Preprocessing consists of sentence splitting, tokenization, POS tagging and lemmatization; the classifier is  a linear SVM. In our experiments (\tabref{ims}), we run IMS with \emph{each feature set by itself} to assess the relative strengths of feature sets (lines 1--7) and on \emph{feature set combinations} to determine which combination is best for WSD (lines 8, 12--15).

IMS implements three standard WSD feature sets: part of speech (POS), surrounding word and local collocation (lines 1--3). 

Let $w$ be an ambiguous word with $k$ senses. The three feature sets on lines 5--7 are based on the AutoExtend embeddings $s^{(j)}$, $1 \leq j \leq k$, of the synsets of $w$ and the centroid $c$ of the sentence in which $w$ occurs. The centroid is simply the sum of all word2vec vectors of the words in the sentence, excluding stop words.

The \textbf{S-cosine} feature set consists of the $k$ cosines of
centroid and synset vectors:
\[
<\cos(c,s^{(1)}),\cos(c,s^{(2)}),\ldots,\cos(c,s^{(k)})>
\]

The \textbf{S-product} feature set consists of the $nk$ element-wise products of centroid and synset vectors:
\[
<c_1 s^{(1)}_1,\ldots,c_n s^{(1)}_n,\ldots,c_1 s^{(k)}_1,\ldots,c_n s^{(k)}_n>
\]
where $c_i$ (resp.\ $s^{(j)}_i$) is element $i$ of $c$ (resp.\ $s^{(j)}$). The idea is that we let the SVM estimate how important each dimension is for WSD instead of giving all equal weight as in S-cosine. 

The \textbf{S-raw} feature set simply consists of the $n(k+1)$ elements of centroid and synset vectors:
\[
<c_1,\ldots,c_n,s^{(1)}_1,\ldots,s^{(1)}_n,\ldots,s^{(k)}_1,\ldots,s^{(k)}_n>
\]

Our main goal is to determine if AutoExtend features improve
WSD performance when added to standard WSD features. To make
sure that improvements we get are not solely due to the
power of word2vec, we also investigate a simple word2vec
baseline. For S-product, the AutoExtend feature set that
performs best in the experiment (cf.\ lines 6 and 14), we
test the alternative word2vec-based
\textbf{S$\dnrm{naive}$-product} feature set.
It has the same definition as
S-product except that we replace the synset vectors $s^{(j)}$
with naive synset vectors $z^{(j)}$, defined as the sum of the word2vec vectors of the words that are members of synset $j$.

Lines 1--7 in \tabref{ims} show the performance of each feature set by
itself.  We see that the synset feature sets (lines 5--7)
have a comparable performance to standard feature sets.
S-product is the strongest of them.

Lines 8--16 show the performance of different feature set
combinations. MFS (line 8) is
the most frequent sense baseline. Lines 9\&10 are the winners of Senseval. The standard configuration of IMS (line 11)
uses the three feature sets on lines 1--3 (POS, surrounding
word, local collocation) and achieves an accuracy of
$65.2\%$ on the English lexical sample task of Senseval-2
and $72.3\%$ on Senseval-3.\footnote{\newcite{zhong2010makes} report accuracies of $65.3\%$ / $72.6\%$ for this configuration.} Lines 12--16 add one
additional feature set to the IMS system on line 11; e.g.,
the system on line 14 uses POS, surrounding word, local
collocation and S-product feature sets. The system on line
14 outperforms all previous systems, most of them
significantly. While S-raw performs quite reasonably as a
feature set alone, it hurts the performance
when used as an additional feature set. As this is the
feature set that contains the largest number of features
($n(k+1)$), overfitting is the likely reason. Conversely,
S-cosine only adds $k$ features and therefore may suffer
from underfitting.\textcolor{white}{\symbolfootnote[2]{In \tabref{ims} and \tabref{scws}, results significantly worse than the best (bold) result in each column are marked $\dagger$ for $\alpha = .05$ and $\ddagger$ for $\alpha = .10$ (one-tailed Z-test).}}

We do a grid search (step size .1) for
optimal values of $\alpha$ and $\beta$, optimizing the average score of Senseval-2
and Senseval-3. The best performing feature set combination is
\textbf{S$\dnrm{optimized}$-product} with $\alpha = 0.2$ and
$\beta = 0.5$, with only a small improvement (line 16).

The main result of this experiment is that we achieve an improvement of more than 1\% in WSD performance when using AutoExtend.

\begin{table}
\small
\centering
\begin{tabular}{lrl|lc}
& & \multicolumn{2}{r}{Senseval-2} & Senseval-3 \\\hline\hline
\multirow{7}{*}{\begin{sideways}IMS feature sets\end{sideways}}
&1&POS & 53.6 & 58.0\textcolor{white}{${}^\dagger$} \\
&2&surrounding word & 57.6 & 65.3\textcolor{white}{${}^\dagger$} \\
&3&local collocation & 58.7 & 64.7\textcolor{white}{${}^\dagger$} \\
&4&S$\dnrm{naive}$-product & 56.5 & 62.2\textcolor{white}{${}^\dagger$} \\
&5&S-cosine & 55.5 & 60.5\textcolor{white}{${}^\dagger$} \\
&6&S-product & 58.3 & 64.3\textcolor{white}{${}^\dagger$} \\
&7&S-raw & 56.8 & 63.1\textcolor{white}{${}^\dagger$} \\\hline
\multirow{9}{*}{\begin{sideways}system comparison\end{sideways}}
&8&MFS & 47.6${}^\dagger$ & 55.2${}^\dagger$ \\
&9&Rank 1 system & 64.2${}^\dagger$ & 72.9\textcolor{white}{${}^\dagger$} \\
&10&Rank 2 system & 63.8${}^\dagger$ & 72.6\textcolor{white}{${}^\dagger$} \\
&11&IMS & 65.2${}^\ddagger$ & 72.3${}^\ddagger$ \\
&12&IMS + S{\tiny naive}-prod. & 62.6${}^\dagger$ & 69.4${}^\dagger$ \\
&13&IMS + S-cosine & 65.1${}^\ddagger$ & 72.4${}^\ddagger$ \\
&14&IMS + S-product & \textbf{66.5} & \textbf{73.6}\textcolor{white}{${}^\dagger$} \\
&15&IMS + S-raw & 62.1${}^\dagger$ & 66.8${}^\dagger$ \\\hline
&16&IMS + S{\tiny optimized}-prod. & \textit{66.6} & \textit{73.6}\textcolor{white}{${}^\dagger$} \\
\end{tabular}
\caption{WSD accuracy for different feature sets and
  systems. Best result (excluding line 16) in each column in bold.}
\tablabel{ims}
\end{table}

\subsection{Synset and lexeme similarity}\seclabel{similarity}
We use SCWS \cite{huang2012improving} for the similarity
evaluation. SCWS provides not only isolated words and
corresponding similarity scores, but also a context for each
word. SCWS is based on WordNet, but the information as to
which synset a word in context came from is not
available. However, the dataset is the closest we could find
for sense similarity. Synset and lexeme embeddings are
obtained by running AutoExtend. Based on the results of the
WSD task, we set $\alpha = 0.2$ and $\beta = 0.5$. Lexeme
embeddings are the natural choice for this task as human
subjects are provided with two words and a context for each
and then have to assign a similarity score. But for
completeness, we also run
experiments for synsets. 

For each word, we compute a context
vector $c$ by adding all word vectors of the context,
excluding the test word itself. Following
\newcite{reisinger2010multi}, we compute the lexeme
(resp.\ synset) vector $l$ either as the simple average of
the lexeme (resp.\ synset) vectors $l^{(ij)}$ (method AvgSim,
no dependence on $c$ in this case) or as the average of the
lexeme (resp.\ synset) vectors weighted by cosine similarity
to $c$ (method AvgSimC).

\tabref{scws} shows that AutoExtend lexeme embeddings (line
7) perform better than previous work, including
\cite{huang2012improving} and
\cite{tian2014probabilistic}. Lexeme embeddings perform
better than synset embeddings (lines 7 vs.\ 6), presumably because
using a representation that is specific to the actual word
being judged is more precise than using a representation
that also includes synonyms.

A simple baseline is to use
the underlying word2vec embeddings directly (line 5).  In
this case, there is only one embedding, so there is no
difference between AvgSim and AvgSimC. It is interesting
that even if we do not take the context into account (method
AvgSim) the lexeme embeddings outperform the original word
embeddings. As AvgSim simply adds up all lexemes of a word,
this is equivalent to the constraint we proposed in the
beginning of the paper (\eqref{word_lexeme}). 
Thus,
replacing a word's embedding
by the sum of the embeddings of its senses could generally 
improve the quality of embeddings (cf.\ 
\newcite{huang2012improving} for a similar point).
We will leave a deeper evaluation
of this topic for future work.
\begin{table}
\small
\centering
\begin{tabular}{rl|cc}
& & AvgSim & AvgSimC \\\hline\hline
1 & \newcite{huang2012improving} 	& 62.8${}^\dagger$ 				& 65.7${}^\dagger$ \\
2 & \newcite{tian2014probabilistic}	& --						& 65.4${}^\dagger$ \\
3 & \newcite{neelakantan2014efficient}	& 67.2\textcolor{white}{${}^\dagger$} 		& 69.3\textcolor{white}{${}^\dagger$} \\
4 & \newcite{chen2014unified} 		& 66.2${}^\dagger$				& 68.9\textcolor{white}{${}^\dagger$} \\\hline
5 & words (word2vec)			& 66.6${}^\ddagger$				& 66.6${}^\dagger$ \\
6 & synsets 				& 62.6${}^\dagger$				& 63.7${}^\dagger$ \\
7 & lexemes 				& \textbf{68.9}\textcolor{white}{${}^\dagger$}	& \textbf{69.8}\textcolor{white}{${}^\dagger$} 
\end{tabular}
\caption{Spearman correlation ($\rho \times 100$) on
  SCWS. Best result per column in bold.}
\tablabel{scws}
\end{table}

\section{Analysis}\seclabel{analysis}
We first look at the impact of the parameters $\alpha$,
$\beta$ (\secref{implementation}) that control
the weighting of synset constraints vs lexeme constraints vs
WN relation constraints. We investigate the impact for three
different tasks. \textbf{WSD-alone:} accuracy of IMS
(average of Senseval-2 and Senseval-3) if only S-product
is used as a feature set (line 6 in
\tabref{ims}). \textbf{WSD-additional:} accuracy of IMS
(average of Senseval-2 and Senseval-3) if S-product is
used together with the feature sets POS, surrounding word and
local collocation (line 14 in \tabref{ims}). \textbf{SCWS:}
Spearman correlation on SCWS (line 7 in \tabref{scws}).

For WSD-alone (\figref{parameters}, center), the best
performing weightings (red) all have high weights for WN
relations and are therefore at the top of triangle. Thus, WN
relations are very important for WSD-alone and adding more
weight to the synset and lexeme constraints does not
help. However, all three constraints are important in
WSD-additional: the red area is in the middle (corresponding
to nonzero weights for all three constraints) in the left
panel of \figref{parameters}. Apparently, strongly weighted
lexeme and synset constraints enable learning of
representations that in their interaction with standard WSD
feature sets like local collocation increase WSD performance.
For SCWS (right panel), we should not put too much weight on WN relations
as they artificially bring related, but not similar lexemes
together. So the maximum for this task is located in the
lower part of the triangle.

The main result of this analysis is that
AutoExtend never achieves its maximum performance when using
only one set of constraints.
All three constraints are important -- synset,
lexeme and WN relation constraints -- with different weights
for different applications.

\begin{figure*}
\centering
\includegraphics[width=16cm]{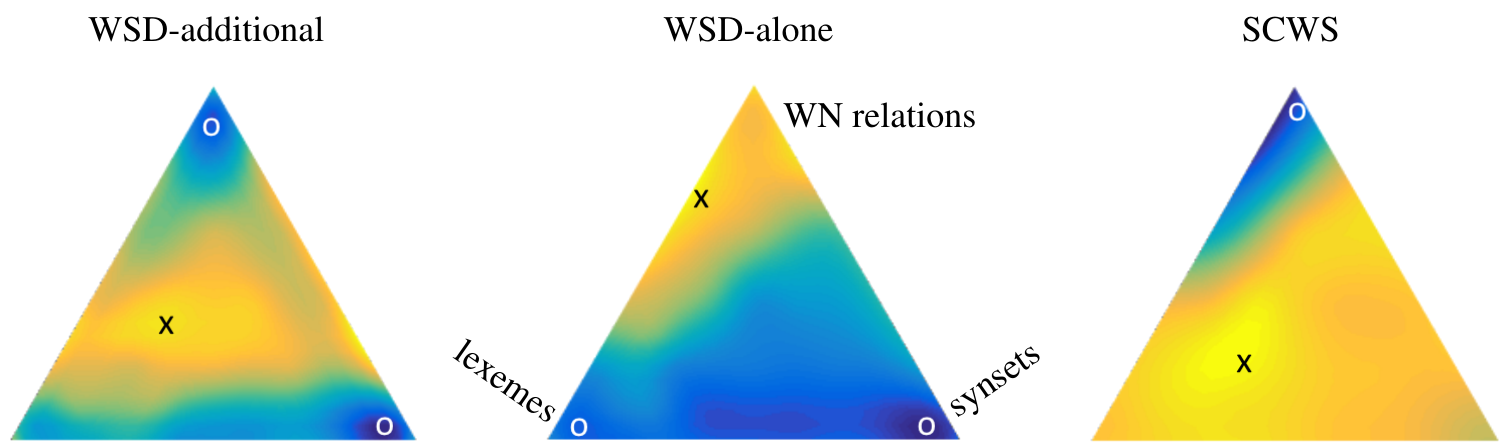}
 \caption{Performance of different weightings of the three
   constraints (WN relations:top, lexemes:left,
   synsets:right) on the three tasks WSD-additional,
   WSD-alone and SCWS.  ``x'' indicates the  maximum; ``o''
   indicates
a local minimum.}
 \figlabel{parameters}
\end{figure*}

We also analyzed the impact of the four different WN
relations (see \tabref{relations}) on performance. In
\tabref{ims} and \tabref{scws}, all four WN relations are
used together. We found that any combination of three
relation types performs worse than using all
four together. A comparison of different relations must be done
carefully as they differ in the POS they affect and in
quantity (see \tabref{relations}). In general, relation
types with more relations outperformed relation types with
fewer relations.

Finally, the relative weighting of $l^{(i,j)}$ and
$\overline{l}^{(i,j)}$ when computing lexeme embeddings is also a
parameter that can be tuned. We use simple averaging
($\theta=0.5$) for all experiments reported in this paper.
We found only small changes in performance  for $0.2
\leq \theta \leq 0.8$.

\section{Resources other than WordNet}\seclabel{extending}
AutoExtend is broadly applicable to lexical and
knowledge resources that have certain properties.
While we only run experiments with WordNet in this paper, we
will briefly address other resources.  For \textit{Freebase}
\cite{bollacker2008freebase}, we could replace the synsets
with Freebase entities. Each entity has several aliases,
e.g. Barack Obama, President Obama, Obama. The role of words
in WordNet would correspond to these aliases in
Freebase. This will give us the synset constraint, as well
as the lexeme constraint of the system. Relations are given
by Freebase types; e.g., we can add a constraint that entity
embeddings of the type "President of the US" should be
similar.

To explorer multilingual word embeddings we require the word embeddings of different languages to live in the same vector space, which can easily be achieved by training a transformation matrix $L$ between two languages using known translations \cite{mikolov2013exploiting}. Let $X$ be a matrix where each row is a word embedding in language 1 and $Y$ a matrix where each row is a word embedding in language 2. For each row the words of $X$ and $Y$ are a translation of each other. We then want to minimize the following objective:
\begin{equation}
 \underset{L}{\operatorname{argmin}} \|LX - Y\|
\end{equation}
We can use a gradient descent to solve this but a matrix inversion will run faster. The matrix $L$ is given by:
\begin{equation}
 L = (X^T * X)^{-1}  (X^T * Y)
\end{equation}
The matrix $L$ can be used to transform unknown embeddings
into the new vector space, which enables us to use a
multilingual WordNet like \textit{BabelNet}
\cite{navigli2010babelnet} to compute synset embeddings. We
can add cross-linguistic relationships to our model,
e.g., aligning German and English synset embeddings of the
same concept.

\section{Related Work}\seclabel{related}
\newcite{rumelhart1988learning} introduced distributed word representations, usually called word embeddings today. There has been a resurgence of work on them recently (e.g., \newcite{bengio2003neural} \newcite{mnih2007three}, \newcite{collobert2011natural}, \newcite{mikolov2013efficient}, \newcite{pennington2014glove}). These models produce only a single embedding for each word. All of them can be used  as input for AutoExtend.

There are several approaches to finding embeddings for
senses, variously called meaning, sense and multiple word
embeddings. \newcite{schutze1998automatic} created sense
representations by clustering context representations
derived from co-occurrence. The representation of a sense is
simply the centroid of its cluster. \newcite{huang2012improving}
improved this by learning single-prototype embeddings before
performing word sense discrimination on
them. \newcite{bordes2011learning} created similarity
measures for relations in WordNet and Freebase to learn
entity embeddings. An energy based model was proposed by
\newcite{bordes2012joint} to create disambiguated meaning
embeddings and \newcite{neelakantan2014efficient} and
\newcite{tian2014probabilistic} extended the Skip-gram model
\cite{mikolov2013efficient} to learn multiple word
embeddings. While these embeddings can correspond to
different word senses, there is no clear mapping between
them and a lexical resource like
WordNet. \newcite{chen2014unified} also modified word2vec
to learn sense embeddings, each
corresponding to a WordNet synset. They use glosses to
initialize sense embedding, which in turn can be used
for WSD. The sense disambiguated
data can again be used to improve sense embeddings.

This prior work needs a training step to learn embeddings. In contrast, we can ``AutoExtend'' any set of given word embeddings -- without (re)training them.

There is only little work on taking existing word embeddings and producing embeddings in the same space. \newcite{labutov2013re} tuned existing word embeddings in supervised training, not to create new embeddings for senses or entities, but to get better predictive performance on a task while not changing the space of embeddings.

Lexical resources have also been used to improve word embeddings. In the Relation Constrained Model, \newcite{yu2014} use word2vec to learn embeddings that are optimized to predict a related word in the resource, with good evaluation results. \newcite{bian2014knowledge} used not only semantic, but also morphological and syntactic knowledge to compute more effective word embeddings.

Another interesting approach to create sense specific word embeddings uses bilingual resources \cite{guo2014learning}. The downside of this approach is that parallel data is needed. 

We used the SCWS dataset for the word similarity task, as it provides a context. Other frequently used datasets are WordSim-353 \cite{finkelstein2001placing} or MEN \cite{bruni2014multimodal}.

And while we use  cosine to compute  similarity between
synsets, there are also a lot of similarity measures
that only rely on a given resource, mostly WordNet. These
measures are often functions that depend on the provided
information like gloss or the topology like
shortest-path. Examples include \cite{wu1994verbs} and
\cite{leacock1998combining}; \newcite{blanchard2005typology}
give a good overview.

\section{Conclusion}
We presented AutoExtend, a flexible method to learn synset and lexeme
embeddings from word embeddings. It is completely general
and can be used for any other set of embeddings and for any
other resource that imposes constraints of a certain type on
the relationship between words and other data types. Our
experimental results show that AutoExtend achieves 
state-of-the-art performance on word similarity and
word sense disambiguation.
Along with this paper, we will publish
AutoExtend for extending word embeddings to other data
types; the lexeme and synset embeddings used in the
experiments; and the code needed to
replicate our WSD evaluation\footnote{\url{http://cistern.cis.lmu.de/}}.

\section*{Acknowledgments}
This work was partially funded by
Deutsche Forschungsgemeinschaft (DFG SCHU 2246/2-2). We are
grateful to Christiane Fellbaum for discussions leading up
to this paper and to the anonymous reviewers for their comments.

\nocite{*}

\bibliography{synset_autoencoder}
\bibliographystyle{acl}

\end{document}